\documentclass[10pt,journal,compsoc]{IEEEtran}
\ifCLASSOPTIONcompsoc
  \usepackage[nocompress]{cite}
\else
  \usepackage{cite}
\fi
\ifCLASSINFOpdf
\else
\fi

\usepackage{epsfig}
\usepackage{graphicx}
\usepackage{amsmath}
\usepackage{amssymb}
\usepackage{bm}
\usepackage{multirow}
\usepackage{booktabs}
\usepackage{enumerate}
\usepackage{hyperref}
\usepackage{changes}
\usepackage{float}
\hypersetup{colorlinks}  
\usepackage{algorithm}
\usepackage{algorithmicx}
\usepackage{algpseudocode}

\algnewcommand\algorithmicinput{\textbf{Input:}}
\algnewcommand\INPUT{\item[\algorithmicinput]}
\algnewcommand\algorithmicoutput{\textbf{Output:}}
\algnewcommand\OUTPUT{\item[\algorithmicoutput]}

\begin{document}
%
\title{Unsupervised Domain Adaptive Person Re-Identification via \\ Human Learning Imitation}
%
%
%
%



\author{Yang Peng *,
        Ping Liu {*},
        Yawei Luo,
        Pan Zhou,
        Zichuan Xu,
        and Jingen Liu
\IEEEcompsocitemizethanks{\IEEEcompsocthanksitem P. Liu is with Center for Frontier AI Research, Agency for Science, Technology, and Research, Singapore. (Email: pino.pingliu@gmail.com) 
\IEEEcompsocthanksitem Y. Peng, P. Zhou are with Hubei Engineering Research Center
on Big Data Security, School of Cyber Science and Engineering, Huazhong University of Science and Technology, Wuhan, Hubei, China.  (Email: yang\_peng@hust.edu.cn, panzhou@hust.edu.cn)
\IEEEcompsocthanksitem Y. Luo is with College of Computer Science and Technology, Zhejiang University, Hangzhou, Zhejiang, China. (Email: yaweiluo329@gmail.com)
\IEEEcompsocthanksitem Z. Xu is with Key Laboratory for Ubiquitous Network and Service
Software of Liaoning Province, School of Software, Dalian University of Technology, Dalian, Liaoning, China. (Email: z.xu@dlut.edu.cn)


\IEEEcompsocthanksitem * means co-first author.
\IEEEcompsocthanksitem P. Liu and P. Zhou are  co-corresponding authors.
}

\thanks{Manuscript received April 19, 2005; revised August 26, 2015.}}

%
%

\markboth{Journal of \LaTeX\ Class Files,~Vol.~14, No.~8, August~2015}%
{Shell \MakeLowercase{\textit{et al.}}: Bare Demo of IEEEtran.cls for Computer Society Journals}
%



\IEEEtitleabstractindextext{%
\begin{abstract}
Unsupervised domain adaptive person re-identification has received significant attention due to its high practical value. In past years, by following the clustering and finetuning paradigm, researchers propose to utilize the teacher-student framework in their methods to decrease the domain gap between different person re-identification datasets. Inspired by recent teacher-student framework based methods, which try to mimic the human learning process either by making the student directly copy behavior from the teacher or selecting reliable learning materials, we propose to conduct further exploration to imitate the human learning process from different aspects, \textit{i.e.}, adaptively updating learning materials, selectively imitating teacher behaviors, and analyzing learning materials structures. The explored three components, collaborate together to constitute a new method for unsupervised domain adaptive person re-identification, which is called Human Learning Imitation framework. The experimental results on three benchmark datasets demonstrate the efficacy of our proposed method.

\end{abstract}

}

\maketitle

\IEEEdisplaynontitleabstractindextext

%
\IEEEpeerreviewmaketitle

\IEEEraisesectionheading{\section{Introduction}\label{sec:intro}}

\begin{figure}[t]
  \centering
   \includegraphics[width=1.0\linewidth]{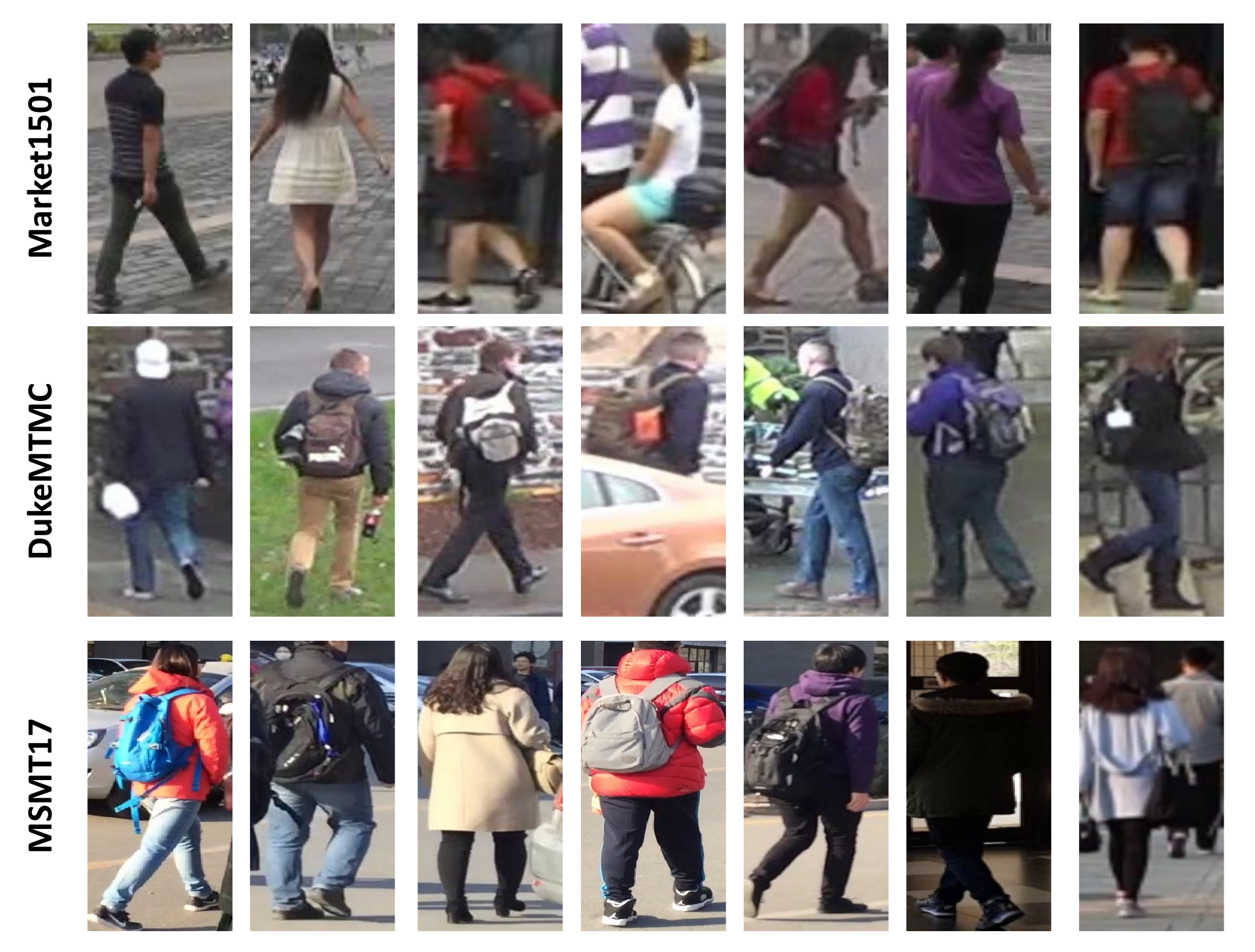}

   \caption{Illustration of domain gap between different person re-ID datasets, \textit{i.e.}, Market1501 \cite{zheng2015scalable_iccv2015}, DukeMTMC \cite{ristani2016performance_eccv2016}, and MSMT17 \cite{wei2018person_cvpr2018}. Due to the different capture circumstances, \textit{e.g.}, weather conditions, countries, and etc, there are significant distribution differences between those datasets.}
   \label{fig:domaingap}
\end{figure}

Person re-identification (re-ID) \cite{9336268_pami2021} is a task to retrieve images of the same person from different image sets. It has received great attention due to its wide applications, \textit{e.g.}, behavior analysis, elderly care, and etc. With the development of deep learning \cite{luo2020adversarial_nips2020,9372870_pami2021}, the performance of person re-ID has achieved significant progresses \cite{Zheng_2017_ICCV, ge2018fd_nips2018, miao2019pose_iccv2019,Zheng_2019_CVPR,lin2019improving_pr2019,9068282_tcsvt2021}. However, those prior works assume that the training set and testing set are with the same statistical distribution, which does not always hold true in real scenarios. As shown in Fig. \ref{fig:domaingap}, in practical cases,  the testing set (target domain) images and training set (source domain) images might be different because of various reasons, \textit{e.g.}, captured by different camera systems/in different weather conditions/from different views. Directly applying a person re-ID model to a new domain might meet a drastic performance drop. To solve this problem, unsupervised domain adaptive (UDA) person re-ID \cite{fan2018unsupervised_tomm2018,8485427_tip2019,ge2020mutual_iclr2020, ge2020improved,Zhang_2021_CVPR} is proposed to deal the domain shift problem in person re-ID.          

Most of recent UDA methods for person re-ID, such as \cite{ge2020mutual_iclr2020, ge2020improved,Zhang_2021_CVPR}, follow the paradigm proposed by Fan \textit{et al.} \cite{fan2018unsupervised_tomm2018}. In \cite{fan2018unsupervised_tomm2018}, Fan \textit{et al.} propose to utilize clustering to generate pseudo labels for unannotated target domain person images with clustering algorithms and refine network based on the generated pseudo labels. The clustering and network finetuning are conducted alternatively until no improvement is observed. Following this paradigm, the performance drop issue in UDA person re-ID was relieved to some extent. However, a further performance improvement is still hindered due to a few reasons, \textit{e.g.}, the noise in the generated pseudo labels. As pointed out in \cite{ge2020mutual_iclr2020}, the quality of generated pseudo labels has key influences on the final performance.

To address the problem of noisy pseudo labels, Ge \textit{et al.} \cite{ge2020mutual_iclr2020} propose an effective unsupervised teacher-student framework, mimicking the human studying process to refine pseudo labels in an iterative manner. The method \cite{ge2020mutual_iclr2020}, which is named Mutual Mean-Teaching (MMT), utilizes a temporal average model to provide reliable supervision signals. The quality of the generated supervision is updated as the model keeps training. When training finishes, the updated network finally captures the target domain distribution, and therefore its generalization ability improves significantly. MMT \cite{ge2020mutual_iclr2020}, and the following works \cite{yang2020asymmetric_aaai2020,chen2021enhancing_wacv2021},  benefit from imitating the human learning process and show significant improvement in UDA person re-ID. Specifically,  \cite{ge2020mutual_iclr2020} keeps updating the teacher network, acting as a better teacher to provide reliable guidance during the learning process; \cite{yang2020asymmetric_aaai2020} designs an asymmetric co-teaching framework to select reliable training data. The achievement of \cite{ge2020mutual_iclr2020,yang2020asymmetric_aaai2020,chen2021enhancing_wacv2021} arouses our curiosity: is it possible to further improve UDA person re-ID performance, by exploring different ways to mimic other aspects in human learning process?

To this end, we conduct an exploration to imitate the human learning process from different manners for improving UDA person re-ID performance. Unlike prior works \cite{ge2020mutual_iclr2020,yang2020asymmetric_aaai2020} focusing on mimicking only one aspect in the human learning process, \textit{e.g.}, providing better teachers \cite{ge2020mutual_iclr2020} or selecting more reliable learning materials \cite{yang2020asymmetric_aaai2020}, we believe that there are more potentialities worth exploiting from the different aspects of human learning process. In this work, we propose to tap more potentials by imitating human learning from three aspects, \textit{i.e.}, updating learning materials adaptively, imitating teacher behavior selectively, analyzing learning materials structurally. Specifically, by adaptively updating the learning materials, the student and teacher are provided more difficult cases, and forced to improve themselves to distill information from the updated learning materials; by imitating teacher behavior in a selective manner, \textit{i.e.}, follow the same action when the teacher makes a correct prediction, but contravene the teacher when an incorrect prediction happens, the student can update itself to an accurate direction; by analyzing the relation between learning materials rather than treating them individually, more structural knowledge can be distilled to provide convincing supervision. We call our proposed method as Human Learning Imitation (HLI) and test it on three benchmark datasets for UDA person re-ID.  

In summary, our contributions are listed as follows:
\begin{itemize}
  \item  We conduct exploration about how to mimic the human learning process for improving the UDA person re-ID performance. Compared to previous works, we propose to imitate the human learning process from three different aspects, \textit{i.e.}, updating learning materials adaptively, imitating teacher behavior selectively, and analyzing learning materials structurally.
  \item We conduct extensive experiments on three benchmark datasets to demonstrate the efficacy of our method. The experimental results show that our proposed produces highly competitive performance in UDA person re-ID compared to prior works.
\end{itemize}

\section{Related Work}
\label{sec:relatedwork}
In  this  section,  we  will  elaborate  on  previous  works  that are  the  most  related  to  our  work,  including unsupervised domain adaptive person re-ID and teacher-student framework.

\subsection{Unsupervised Domain Adaptive Person Re-ID}
Person re-id \cite{9336268_pami2021} aims to conduct person retrieval from different image sets, which might be captured by different cameras. In real scenarios, a person re-id model might be trained and tested on two different domains (source/target) with different statistical distributions. To handle the domain gap issues in this scenario, unsupervised domain adaptive person re-identification methods are proposed. Most prior UDA person re-ID methods can be categorized into two groups. One is style transfer based methods  \cite{wei2018person_cvpr2018,deng2018image_cvpr2018} and the other one is clustering based methods  \cite{fan2018unsupervised_tomm2018,ge2020mutual_iclr2020,ge2020improved,Zhang_2021_CVPR}.

Style transfer based methods try to align distribution between domains for minimizing the domain gap. The alignment can be conducted on image-level \cite{wei2018person_cvpr2018,deng2018image_cvpr2018}, which utilize Cycle-GAN \cite{zhu2017unpaired_iccv2017} to translate the foreground of source images to target domain. The shortage of the style transfer methods, such as \cite{wei2018person_cvpr2018,deng2018image_cvpr2018}, is that they might change the appearance or attribute of the foreground of source images, changing the identity of translated images. 

Clustering based methods \cite{fan2018unsupervised_tomm2018,ge2020mutual_iclr2020,ge2020improved,Zhang_2021_CVPR} are another kind of domain adaptive person re-ID methods. In general, those methods use clustering techniques to group unlabeled target domain images for generating pseudo labels. The generated pseudo labels are utilized to finetune a pretrained model. The pseudo labels generation via clustering and model finetuning are conducted in an alternative manner until no further improvement observed. One of the keys of the clustering based methods is the quality of generated pseudo labels, which, unfortunately, can not be avoided due to the clustering. Utilizing the sample with incorrect pseudo labels have a negative impact on final performance of domain adaptive person re-ID. To handle the noise label issue, Ge \textit{et al.} \cite{ge2020mutual_iclr2020} and the following works \cite{yang2020asymmetric_aaai2020,chen2021enhancing_wacv2021}  propose to utilize teacher-student framework to refine the label quality, which is mimicking one aspect of the human learning process: student follows teacher behavior during learning, or selecting reliable learning materials.

\subsection{Teacher-Student Framework}
Teacher-student framework is also known as knowledge distillation \cite{9340578_pami2020}, which has been widely applied in various computer vision and machine learning tasks, \textit{e.g.}, model compression \cite{luo2016face_aaai2016, wang2019private_aaai2019}, knowledge transfer \cite{xu2020knowledge_eccv2020, liu2021point_cyber2021}, and etc. Generally speaking, the common characteristic of teacher-student framework is to transfer knowledge from one network, \textit{i.e.}, the teacher, to another network, \textit{i.e.}, the student. The difference is ``what" to transfer between them.

In previous works on teacher-student framework, the knowledge transferred between networks can be \textit{logits},  \textit{intermediate features}, and \textit{attention maps}.
Knowledge from logits is also called softened labels, which is proposed by Hinton \textit{et al.} \cite{hinton2015distilling}. Given the same data, the logits from the student and the logits from the teacher are requested to be similar by minimizing a predefined loss function. Knowledge transfer from intermediate features \cite{aguilar2020knowledge_aaai2020} propose to link the behavior in the intermediate layers between the teacher and student. In knowledge transfer from attention maps \cite{komodakis2017paying_iclr2017}, Zagoruyko \textit{et al.} propose to improve the student performance by mimicking the attention maps from the teacher.

Teacher-student framework has been utilized to improve the performance in UDA person re-ID \cite{ge2020mutual_iclr2020,yang2020asymmetric_aaai2020,chen2021enhancing_wacv2021}. They try to imitate the human learning process from a solo aspect: \cite{ge2020mutual_iclr2020} focuses on providing a better teacher by utilizing a temporal average strategy; \cite{chen2021enhancing_wacv2021} aims to select reliable learning materials for training the student. Our work, from a different perspective, explores more ways to imitate the human learning process for improving the UDA person re-ID performance.
\section{Methodology}

\label{sec:method}
  \subsection{Problem Definition and Preliminaries}\label{sub1}
   
  Unsupervised domain adaptive person re-identification targets to train a model based on annotated source domain data, and achieve high accuracy on target domain. The source domain data is denoted as $\mathbb{D}_{s}=\left\{\left.\left(\boldsymbol{x}_{i}^{s}, \boldsymbol{y}_{i}^{s}\right)\right|_{i=1} ^{N_{s}}\right\}$, in which $\boldsymbol{x}_{i}^{s}$ and $\boldsymbol{y}_{i}^{s}$ denote $i$-th sample and the corresponding identity label in source domain, $N_{s}$ is the source image number.  The target domain data is denoted as $\mathbb{D}_{t}=\left\{\left.\left(\boldsymbol{x}_{i}^{t}\right)\right|_{i=1} ^{N_{t}}\right\}$, in which $\boldsymbol{x}_{i}^{t}$ denotes $i$-th target domain image.
   
  Recent works on UDA person re-ID, such as \cite{fu2019self_iccv2019, ge2020mutual_iclr2020}, train the model in a three-stage manner: (1) pretrain a model $F(.; \theta)$ based on annotated source domain data $\boldsymbol{x}_{*}^{s}$, where $\theta$ denotes the network parameters; (2) use the trained network $F(.; \theta)$ to generate pseudo label $\boldsymbol{y}_{*}^{t}$ for target domain data $\boldsymbol{x}_{*}^{t}$ based on clustering, where the clustering number $M_{t}$ is pre-defined; (3) refine the model $F(.; \theta)$ by optimizing with respect to the generated pseudo labels for target data. The stage (2) and (3) are conducted alternatively until no improvement observed. The pseudo label quality in stage (2) has a direct influence on the final performance, and therefore, recent works, such as \cite{ge2020mutual_iclr2020,yang2020asymmetric_aaai2020,chen2021enhancing_wacv2021}, propose to utilize teacher-student framework. 
  
  In this work, we conduct exploration for imitating human learning process from new aspects, for improving the UDA person re-ID performance.
   
  \subsection{Human Learning Imitation}
   
\begin{figure*}[t]
  \centering
   \includegraphics[width=0.6\linewidth]{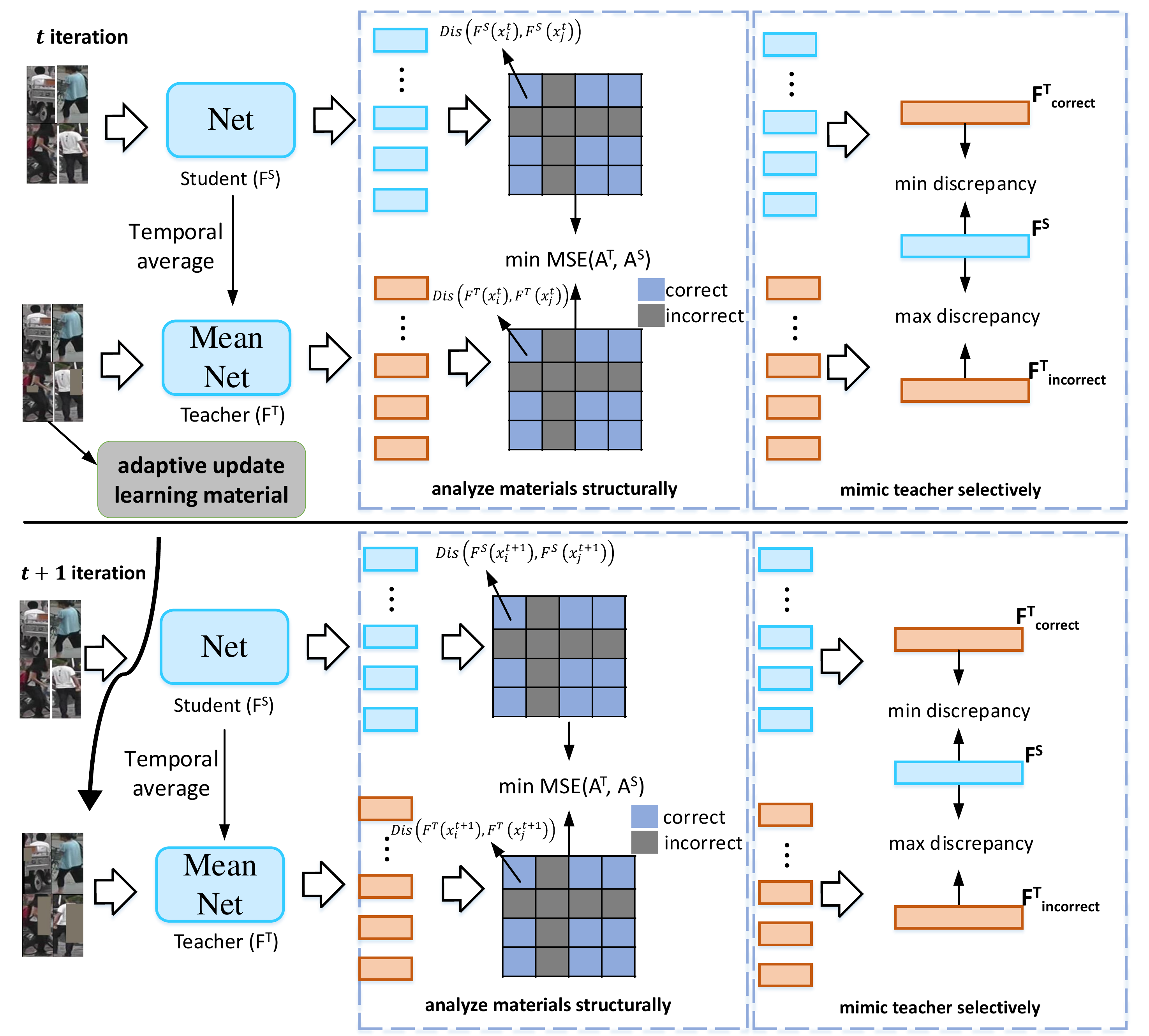}

   \caption{The pipeline of our proposed Human Learning Imitation (HLI) framework. Our method taps more potentials by imitating human learning from three aspects, \textit{i.e.}, updating learning materials adaptively, imitating teacher behavior selectively, analyzing learning materials structurally. Specifically, by adaptively updating the learning materials, the student and teacher are provided more difficult cases, and forced to improve themselves to distill information from the updated learning materials; by imitating teacher behavior in a selective manner, \textit{i.e.}, follow the same action when the teacher makes a correct prediction, but contravene the teacher when an incorrect prediction happens, the student can update itself to an accurate direction; by analyzing the relation between learning materials rather than treating them individually, more structural knowledge can be distilled to provide convincing supervision.}
   \label{fig:pipeline}
\end{figure*}

  Based on the observations, there are a few key factors in the human learning process: (1) providing learning materials based on the learning capability of students; (2) selectively mimic behaviors of the teacher during studying; (3) learning the materials systematically and structurally by considering their internal relations, \textit{i.e.}, similarities and differences, in a systematical and structural manner. Inspired by those factors, we conduct exploration about how to improve unsupervised domain adaptive person re-identification by imitating human learning process from three different but related aspects, \textit{i.e.}, Adaptively Updating Learning Materials (\textbf{AULM}), Imitating Teacher Behavior Selectively (\textbf{ITBS}), Analyzing Learning Materials Structurally (\textbf{ALMS}).  We call our method as Human Learning Imitation (\textbf{HLI}), which is shown in Fig. \ref{fig:pipeline}. All the details are illustrated in the following subsections. 
   \vspace{-0.4cm}

     \subsubsection{Adaptive Updating Learning Materials }
  In human learning process, the learning materials are designed based on the learning capability of the students. For example, when students are promoted to a higher grade, they are issued new textbooks with more complex content. Inspired by this, we propose to keep updating learning materials in an adaptive manner, \textit{i.e.},  based on the network status. The updated learning materials is generated by simultaneously considering the student status and recently used learning material contents, and therefore suitable for network improvement.
  
  
  Update learning materials adaptively is a design principle, and therefore, any strategy matches this principle can be utilized. In this work, we utilize a simple yet effective method to update the learning materials adaptively. For $i$-th sample $\boldsymbol{x}_{i}^{}$, we use the current network $F(.; \theta)$ to locate the most informative point in it, which is denoted as $(P_x^{},P_y^{})$ . Specifically, the network $F(.; \theta)$ trained in the current epoch is used to generate the heat map \cite{zhou2016learning_cvpr2016} of $\boldsymbol{x}_{i}^{}$, which is denoted as $Cam(x_i^{})$. The most informative position for $\boldsymbol{x}_{i}^{}$ is defined as follows:
  \begin{equation}
      (P_x^{},P_y^{}) = argmax(Cam(\boldsymbol{x}_{i}^{}))
  \label{equ1}
  \end{equation}
  where the point with the maximum value in $Cam(\boldsymbol{x}_{i}^{})$, \textit{i.e.},$ (P_x^{},P_y^{})$,  corresponds the the most informative position for the image $\boldsymbol{x}_{i}^{}$. As pointed out by \cite{zhou2016learning_cvpr2016}, the heat map represents the current cognitive situation of the network, in which the larger value indicates greater influence on the final result. Based on this observation, the updated learning material is generated by erasing a region with the size of $w \times h$ centered at the point $(P_x^{},P_y^{})$. The updated learning materials are utilized to update the network, which, conversely, benefits to update learning materials.
  
  The updated materials are highly adaptive due to two reasons: (1) for different samples, the most informative positions are usually different; (2) given different networks , for example, networks trained in different epochs, the most informative points for the same sample are also different due to different network status.
  
  \subsubsection{Imitating Teacher Behavior Selectively}
  Most prior teacher-student framework based methods directly make the student mimic the behavior of the teacher, even when the teacher make an incorrect prediction. The knowledge in wrong predictions, also known as dark knowledge, is considered to regularize the student network to extract more meaningful knowledge \cite{yun2020regularizing_cvpr2020}.
  
 We argue that directly mimicking the behaviors of the teacher might not be the best choice. On the one hand, as pointed out by \cite{huang2021revisiting_cvpr2021}, directly making student mimic the teacher might limit  student capability during learning; on the other hand, based on intuition, learning from mistakes might be necessary and beneficial, but mechanically repeating mistakes is not.
  
  Based on this, we propose to mimic teacher behavior in a \textit{selective} manner: we make the student follow the teacher behavior if the teacher makes correct predictions; when the teacher makes incorrect predictions, we push the student not to follow but: (1) explore a different, \textit{e.g.}, opposite action from the teacher; (2) take advantage of the teacher supervision if the teacher makes correct predictions when dealing with similar learning materials.
  
  
For the following discussion, we denote the features from the teacher network as $F_{T} (.; \theta)$, the features from the student network as $F_S (.; \theta)$.  Based on the classification result of teacher network (correct/incorrect), we divided  $F_T (.; \theta)$ into $F_{T}^{cor} (.;\theta)$ and $F_{T}^{inc} (.; \theta)$ . Unlike previous works directly minimizing the discrepancy between $F_S (.; \theta)$  and  $F_{T} (.; \theta)$, we make the student mimic the teacher selectively. Specifically, we minimize the difference between $F_S (.; \theta)$  and $F_{T}^{cor} (.;\theta)$ by optimizing the \textit{mimic} loss $\mathcal{L}_{mim}$, making the student mimic the behaviors of the teacher only when the teacher is correct. $\mathcal{L}_{mim}$ is formulated as follows:
  \begin{equation}
      \mathcal{L}_{mim}=\mathcal{\textbf{MSE}}( F_S (.; \theta),F_{T}^{cor} (.;\theta))
  \label{eq: lmim}
  \end{equation}
where \textbf{MSE} denotes the Mean Square Error loss.
  
When the teacher makes incorrect predictions, we encourage the student to exploit diverse knowledge rather than directly copy the mistakes from the teacher. For this purpose, we design the \textit{exploitation} loss $\mathcal{L}_{exp}$ by considering two aspects: one one hand, we maximize the difference between $F_S (.; \theta)$ and $F_{T}^{inc} (.;\theta)$; on the other hand, if the teacher makes the correct prediction with other images of the same person, we utilize the average response on those images from the teacher, which is denoted as $F_{T}^{cor'} (.;\theta)$, to guide the student. The formulation is defined as follows:

  

 \begin{align}
      \mathcal{L}_{exp}&=-\textbf{MSE}(F_S(.; \theta),F_T^{inc}(.; \theta))\notag \\
      &+\textbf{MSE}(F_S(.; \theta), F_T^{cor'}(.; \theta))
    \label{equ3}
\end{align}

The formulations of $F_{T}^{inc}(.;\theta)$ and $F_{T}^{cor'}(.;\theta)$ are defined as follows: we assume that there are $N$ images for the $i$-th identity in the given batch data. In the $N$ images with the same identity, we assume that the teacher makes incorrect predictions for $M$ images, where $M < N$, correct predictions for the left $N-M$ images. For each sample in the $N$ images, \textit{i.e.}, $\bm{x_i}, 1 \leq i \leq N$, we calculate $F_{T}^{inc}(.;\theta)$ and $F_{T}^{cor'}(.;\theta)$ as follows: $F_{T}^{inc}(\bm{x_i};\theta)$ = $F_{T}^{}(\bm{x_i};\theta)$, where $i \in M$; $F_{T}^{cor'}(\bm{x_i};\theta) = \frac{\sum_{1}^{N-M}F_{T}^{}(\bm{x_j};\theta)}{N-M}$, where $i \in M, j \in N-M$.


The total loss for imitating teacher selectively, denoted as $\mathcal{L}_{imi}$, is the weighted sum of $\mathcal{L}_{mim}$ and $\mathcal{L}_{exp}$, which is formulated as follows:
\begin{equation}
   \mathcal{L}_{imi}=\alpha \mathcal{L}_{mim}+\beta \mathcal{L}_{exp}
\label{equ4}
\end{equation}
where $\alpha$ and $\beta$ are weighting parameters.

\subsubsection{Learning the Materials Structurally}
When studying new materials, it is very common to group the learning materials, \textit{e.g.}, exercises, based on the sharing patterns. For example, in LeetCode, questions are grouped based on the category, \textit{i.e.}, Dynamic Programming, Hash Table, and etc. The structural information conveyed by the similarities and differences between instances, provide more systematical supervision to the student. Inspired by this, we propose to utilize the relationship between person images in the same batch to provide structural knowledge to transfer. Specifically, for all the samples in the same batch, we construct a relation matrix $\textbf{A} \in \mathbb{R}^{N\times N}$, in which $N$ denotes the sample number in a batch. As there is a student network and a teacher network, both of which can be used to calculate a relation matrix, we construct two relation matrices, denoted as  $\textbf{A}^{S}$ and $\textbf{A}^{T}$, respectively.

Each element in the relation matrix $\textbf{A}$, \textit{e.g.}, $\textbf{A}_{i,j}$, denotes the feature discrepancy between sample $\boldsymbol{x}_{i}^{*}$ and $\boldsymbol{x}_{j}^{*}$ \footnote{* denotes s (source) or t (target).}, which is formulated as follows:
\begin{equation}
    \textbf{A}_{i,j}^{*}=\textbf{Dis}(F_{S}(\boldsymbol{x}_{i}^{*}, \theta),F_{S}(\boldsymbol{x}_{i}^{*}, \theta)
\label{equ5}
\end{equation}
where \textbf{Dis} is a metric function, here we choose the normalized inverse Euclidean distance. To transfer the structural knowledge from the teacher to the student, we minimize the discrepancy between the two relation matrices by optimizing the following structure distillation loss functions, $\mathcal{L}_{sd}$:

\begin{equation}
        \mathcal{L}_{sd}=\textbf{MSE} (\textbf{A}_{}^{S}, \textbf{A}_{}^{T})
\label{equ7}
\end{equation}

\subsection{Training Details}
\textbf{Supervised pre-training on the source domain}
Following the paradigm proposed by Fan \textit{et al.} \cite{fan2018unsupervised_tomm2018}, we utilize the source domain image to pretrain a network $F_{S} (,; \theta)$.
\textbf{Clustering and Refinement}
Following MMT\cite{ge2020mutual_iclr2020}, we use the pre-trained network as the student and its temporal averaged network as the teacher $F_{T} (,; \theta)$.The model is trained with the knowledge diversity part and structure distillation part, as well as the updated learning materials. The overall loss function is defined as follows:
\vspace{-0.2cm}

\begin{equation}
\begin{aligned}
    \mathcal{L}(\theta)&=\left(1-\lambda_{i d}^{t}\right) \mathcal{L}_{\mathrm{id}}^{t}(\theta)+\lambda_{i d}^{t} \mathcal{L}_{s i d}^{t}(\theta)+\left(1-\lambda_{t r i}^{t}\right)\\& \mathcal{L}_{t r i}^{t}(\theta)+\lambda_{\mathrm{t r i}}^{t} \mathcal{L}_{s t r i}^{t}(\theta)+\lambda_{imi}^{t} \mathcal{L}_{imi}+\lambda_{sd}^{t} \mathcal{L}_{sd}
\end{aligned}
\label{equ8}
\end{equation}

In Eq. \ref{equ8}, $\mathcal{L}_{\mathrm{id}}^{t}$,  $\mathcal{L}_{s i d}^{t}$, $\mathcal{L}_{t r i}^{t}$, and $ \mathcal{L}_{s t r i}^{t}$ are loss terms proposed in MMT \cite{ge2020mutual_iclr2020}\footnote{Please check the supplementary for details.}, which have been demonstrated effective. To make a fair comparison, we also utilize them in our loss terms. $ \mathcal{L}_{imi}$ and $\mathcal{L}_{sd}$ are formulated in Eq. \ref{eq: lmim} - Eq. \ref{equ7}, $\lambda_{i d}^t,\lambda_{tri}^t, \lambda_{imi}^{t}$and $\lambda_{sd}^t$ are weighted parameters. 


\section{Experimental Results}
\label{section:experiments_results}

To demonstrate the effectiveness of our proposed method, we conduct extensive experiments to make a comparison with prior works. In \ref{subsec:datasets}, we illustrate the dataset details and evaluation metrics. The implementation details are provided in \ref{subsec:implementationdetails}. In \ref{subsec:sotacomparison}, we compare our proposed method with prior state-of-the-art methods on four transfer settings. The ablation study and 
parameter sensitivity are in \ref{subsec:ablation}.


\subsection{Datasets}
\label{subsec:datasets}
To make fair comparisons with prior works, we utilize the same person re-ID datasets, which include Market1501 \cite{zheng2015scalable_iccv2015}, DukeMTMC \cite{ristani2016performance_eccv2016}, and MSMT17 \cite{wei2018person_cvpr2018}. 


\textbf{Market1501} \cite{zheng2015scalable_iccv2015} In Market1501, there are $32,668$ images of $1,501$ person taken by 6 cameras. In the training set there are $12,936$ annotated images with $751$ identities, in the test set there are $19,732$ images with $750$ person.

\textbf{DukeMTMC-ReID} \cite{ristani2016performance_eccv2016} In DukeMTMC-ReID, there are $36, 411$ annotated images containing $1,404$ person. There are $16,522$ images of $702$ person in the training set, $19,889$ images of $702$ person in the testing set. 
The images are captured by $8$ different cameras.

\textbf{MSMT17} \cite{{wei2018person_cvpr2018}} In MSMT17, there are $126,441$ images of $4,101$ person captured by $15$ cameras deployed in a campus. The training set contains $32,621$ images of $1,041$ person. The query set and test set contain $11,659$ and $82,161$ images of $3,060$ person respectively.

To make a fair comparison with prior works, we follow the same set up as \cite{ge2020mutual_iclr2020}, which include:  DukeMTMC-to-Market1501, Market1501-to-DukeMTMC, DukeMTMC-to-MSMT17 and Market1501-to-MSMT17. 

\textbf{Evaluation Metrics} In our experiment, Cumulative Matching Characteristic (CMC) curve and Mean Average Precision (mAP) are used to comprehensively evaluate the performance of our method. The rank-$k$ values in the CMC curve indicate the probability that a query image appear in the top-$k$ positions. mAP is an evaluation metrics frequently used in retrieval tasks.

\begin{table*}[] \tiny
\centering
\resizebox{1.0\linewidth}{!}{
\begin{tabular}{l|cccc|cccc}
\hline
\multicolumn{1}{c}{\multirow{2}{*}{Methods}} & \multicolumn{4}{|c|}{Marke1501→MSMT17} & \multicolumn{4}{c}{DukeMTMC→MSMT17} \\
\cline{2-9}
\multicolumn{1}{c|}{}                         & mAP      & top-1     & top-5     & top-10     & mAP      & top-1     & top-5     & top-10     \\
\hline
SSG \cite{fu2019self_iccv2019}(ICCV'19)                           & 13.2   & 31.6    & -       & 49.6    & 13.3   & 32.2    & -      & 51.2    \\
MMT-500\cite{ge2020mutual_iclr2020}(ICLR'20)(ResNet-50)             & 16.6   & 37.5    & 50.6    & 56.5    & 17.9   & 41.3    & 54.2   & 59.7    \\
MMT-1000\cite{ge2020mutual_iclr2020}(ICLR'20)(ResNet-50)            & 21.6   & 46.1    & 59.8    & 66.1    & 23.5   & 50.0    & 63.6   & 69.2    \\
MMT-1500\cite{ge2020mutual_iclr2020}(ICLR'20)(ResNet-50)            & 22.9   & 49.2    & 63.1    & 68.8    & 23.3   & 50.1    & 63.9   & 69.8    \\
MMT-2000\cite{ge2020mutual_iclr2020}(ICLR'20)(ResNet-50)            & 22.9   & 49.2    & 63.1    & 68.8    & 23.3   & 50.1    & 63.9   & 69.8    \\
MMT-500\cite{ge2020mutual_iclr2020}(ICLR'20)(IBN-ResNet-50)             & 19.6   &43.3    & 56.1    & 61.6    & 23.3   & 50.0    & 62.8   & 68.4    \\
MMT-1000\cite{ge2020mutual_iclr2020}(ICLR'20)(IBN-ResNet-50)             & 26.3  &52.5    & 66.3    & {71.7}    & \underline{29.7}   & \underline{58.8}    & 71.0   & 76.1    \\
MMT-1500\cite{ge2020mutual_iclr2020}(ICLR'20)(IBN-ResNet-50)             & \underline{26.6}  & 54.4    & \underline{67.6}    & \underline{72.9}    & 29.3   & 58.2    & \underline{71.6}   & \underline{76.8}    \\
MMT-2000\cite{ge2020mutual_iclr2020}(ICLR'20)(IBN-ResNet-50)             & 25.1  & 52.7    & 65.9    & 71.3    & 28.1   & 56.8    & 70.8   & 76.0    \\
NRMT \cite{zhao2020unsupervised_eccv2020}(ECCV'20)                          & 19.8   & 43.7    & 56.5    & 62.2    & 20.6   & 45.2    & 57.8   & 63.3    \\
ABMT \cite{chen2021enhancing_wacv2021}(WACV'2021)                        & 23.2   & 49.2    & -       & -       & 26.5   & 54.3    & -      & -       \\
GLT \cite{Zheng_2021_CVPR2021}(CVPR'2021)                         & 26.5   & \underline{56.6}
& 67.5    & 72.0    & 27.7   & 59.5    & 70.1   & 74.2    \\
UNRN \cite{abs-2012-08733_aaai2021}(AAAI'2021)                         & 25.3  & 52.4    & 64.7    & 69.7    & 26.2   & 54.9    & 67.3   & 70.6    \\

\hline
HLI-500(ResNet-50)                            & 21.7   & 47.2    & 61.1    & 66.5    & 23.8   & 52.5    & 64.4   & 69.4    \\
HLI-1000(ResNet-50)                           & 29.4   & 59.1    & 71.7    & 76.8    & 30.1   & 61.0    & 72.9   & 77.5    \\
HLI-1500(ResNet-50)                           & 30.2   & 61.0    & 73.2    & 77.8    & 30.8   & 62.7    & 74.7   & 79.1    \\
HLI-2000(ResNet-50)                           & 28.2   & 59.3    & 72.2    & 76.8    & 28.7   & 60.8    & 73.1   & 77.6    \\
\hline
HLI-500(IBN-ResNet-50)                        & 24.4   & 52.1    & 64.9    & 70.2    & 26.4   & 55.7    & 67.5   & 72.5    \\
HLI-1000(IBN-ResNet-50)                       & {32.9}   & 63.0    & 74.7    & 79.7    & 34.1   & 65.0    & 77.1   & 81.1    \\
HLI-1500(IBN-ResNet-50)                       & \textbf{32.9}   & \textbf{64.4}    & \textbf{76.5}    & \textbf{80.9}    & \textbf{34.8}   & \textbf{66.9}    & \textbf{78.6}   & \textbf{82.4}    \\
HLI-2000(IBN-ResNet-50)                       & 30.2   &61.5    & 74.1    & 78.9    & 31.4   & 63.6    & 75.5   & 79.7    \\
\hline
\hline
\end{tabular}
}
\caption{Performance comparisons to the unsupervised domain adaptive person re-ID state-of-the-art methods transferring from small dataset, \textit{i.e.}, Market1501 \cite{zheng2015scalable_iccv2015}, DukeMTMC \cite{ristani2016performance_eccv2016}, to large dataset, \textit{i.e.}, MSMT17 \cite{wei2018person_cvpr2018}. }
\label{tab:comparison_small_to_large}
\end{table*}

\begin{table*}[] \tiny
\centering
\resizebox{1.0\linewidth}{!}{
\begin{tabular}{l|cccc|cccc}
\hline
\multicolumn{1}{c}{\multirow{2}{*}{Methods}} & \multicolumn{4}{|c|}{DukeMTMC→Market1501} & \multicolumn{4}{c}{Market1501→DukeMTMC} \\
\cline{2-9}
\multicolumn{1}{c|}{}                         & mAP      & top-1     & top-5     & top-10     & mAP      & top-1     & top-5     & top-10     \\
\hline
CFSM \cite{chang2019disjoint_aaai2019}(AAAI'19)                          & 28.3     & 61.2      & -         & -          & 27.3     & 49.8      & -         & -          \\
BUC\cite{lin2019bottom_aaai2019}(AAAI'19)                           & 38.3     & 66.2      & 79.6      & 84.5       & 27.5     & 47.4      & 62.6      & 68.4       \\
ENC\cite{zhong2019invariance_cvpr2019}(CVPR'19)                           & 43.0     & 75.1      & 87.6      & 91.6       & 40.4     & 63.3      & 75.8      & 80.4       \\
PDA-Net\cite{li2019cross_iccv2019}(ICCV'19)                       & 47.6     & 75.2      & 86.3      & 90.2       & 45.1     & 63.2      & 77.0      & 82.5       \\
PCB-PAST \cite{zhang2019self_iccv2019}(ICCV'19)                      & 54.6     & 78.4      & -         & -          & 54.3     & 72.4      & -         & -          \\
SSG \cite{fu2019self_iccv2019}(ICCV'19)                           & 58.3     & 80.0      & 90.0      & 92.4       & 53.4     & 73.0      & 80.6      & 83.2       \\
ACT\cite{yang2020asymmetric_aaai2020}(AAAI'20)                           & 60.6     & 80.5      & -         & -          & 54.5     & 72.4      & -         & -          \\
MPLP+MMCL \cite{wang2020unsupervised_cvpr2020}(CVPR'20)                     & 60.4     & 84.4      & 92.8      & 95.0       & 51.4     & 72.4      & 82.9      & 85.0       \\
AD-Cluster \cite{zhai2020ad_cvpr2020}(CVPR'20)                    & 68.3     & 86.7      & 94.4      & 96.5       & 54.1     & 72.6      & 82.5      & 85.5       \\
MMT-500\cite{ge2020mutual_iclr2020}(ICLR'20)(ResNet-50)             & 71.2     & 87.7      & 94.9      & 96.9       & 63.1     & 76.8      & 88.0      & 92.2       \\
MMT-700\cite{ge2020mutual_iclr2020}(ICLR'20)(ResNet-50)             & 69.0     & 86.8      & 94.6      & 96.9       & 65.1     & 78.0      & 88.8      & 92.5       \\
MMT-900\cite{ge2020mutual_iclr2020}(ICLR'20)(ResNet-50)             & 66.2     & 86.8      & 94.9      & 96.6       & 63.1     & 77.4      & 88.1      & 92.5       \\
MMT-500\cite{ge2020mutual_iclr2020}(ICLR'20)(IBN-ResNet-50)             & 76.5     & 90.9      & 96.4      & 97.9       & 65.7     & 79.3      & 89.1      & 92.4       \\
MMT-700\cite{ge2020mutual_iclr2020}(ICLR'20)(IBN-ResNet-50)             & 74.5     & 91.1      & 96.5      & 98.2       & 68.7     & 81.8      & \underline{91.2}      & 93.4       \\
MMT-900\cite{ge2020mutual_iclr2020}(ICLR'20)(IBN-ResNet-50)             & 72.7     & 91.2      & 96.3      & 98.0       & 67.3     & 80.8      & 90.3      & 93.0       \\
NRMT \cite{zhao2020unsupervised_eccv2020}(ECCV'20)                          & 71.7     & 87.8      & 94.6      & 96.5       & 62.2     & 77.8      & 86.9      & 89.5       \\
SpCL \cite{ge2020self_nips2020}(NeurIPS'20)                   & 76.7     & 90.3      & 96.2      & 97.7       & 68.8     & \underline{82.9}      & 90.1      & 92.5       \\
 MEB-Net \cite{zhai2020multiple_eccv2020}(ECCV'20)                           & 76.0     & 89.9      & 96.0      & 97.5       & 66.1     & 79.6      & 88.3      & 92.2       \\

GLT \cite{Zheng_2021_CVPR2021}(CVPR'2021)                         & \underline{{79.5}}   & \underline{92.2}    & \underline{96.5}    & \underline{97.8}    & \underline{69.2}   & 82.0    & 90.2   & 92.8    \\
UNRN \cite{abs-2012-08733_aaai2021}(AAAI'2021)                         & 78.1  & 91.9    & 96.1    & 97.8    & 69.1   & 82.0    & 90.7   & \underline{93.5}    \\
\hline
HLI-500(ResNet-50)                            & 74.3     & 89.8      & 96.2      & 97.8       & 65.6     & 80.0      & 89.8      & 92.2       \\
HLI-700(ResNet-50)                            & 73.9     & 90.6      & 96.6      & 98.0       & 66.4     & 80.1      & 90.4      & 92.8       \\
HLI-900(ResNet-50)                            & 70.6     & 89.8      & 96.1      & 97.7       & 65.4     & 79.4      & 89.1      & 92.4  
 \\
\hline
HLI-500(IBN-ResNet-50)                            & \textbf{76.5}     & 91.2      & 96.6      & 97.9       & 65.6     & 80.0      & 89.8      & 92.2       \\
HLI-700(IBN-ResNet-50)                            & 75.2     & \textbf{91.6}      & \textbf{96.9}      & 98.0       & \textbf{68.8}     & \textbf{82.5}      & \textbf{90.7}      & \textbf{93.0}       \\
HLI-900(IBN-ResNet-50)                            & 74.1     & 91.6      & 96.9      & \textbf{98.1}       & 67.5     & 81.8      & 90.4      & 93.0 
 \\
\hline
\hline
\end{tabular}
}
\caption{Performance comparisons to the unsupervised domain adaptive person re-ID state-of-the-art methods transferring between Market1501 \cite{zheng2015scalable_iccv2015} and DukeMTMC \cite{ristani2016performance_eccv2016}. }
\label{tab:comparison}
\end{table*}

\subsection{Implementation Details}
\label{subsec:implementationdetails}
\subsubsection{Source domain pre-training }

In the pre-training stage, we use ResNet-50 \cite{he2016deep_cvpr2016} and IBN ResNet-50 \cite{pan2018two_eccv2018} pre-trained on ImageNet \cite{deng2020imagenet_cvpr2009} as the backbone of the network. For the input image , in order to meet the traditional hard batch triple loss, we sample randomly selected $4$ photos for each person, and  resize the image to $256 \times 128$. We set the batch size to $64$.  For each mini batch sample, the network parameter was updated with weight factor $\lambda_s=1$. The initial learning rate is set to $0.00035$ and will be decayed to 1/10  after 40th and 70th epoch. We choose the Adam optimizer \cite{kingma2018method_iclr2015} with weight delay 0.0005.


\subsubsection{Unsupervised domain adaptive training}
 During the training process, the network parameters are updated by optimizing Eq. \ref{equ8} with the loss weights $\lambda_{id}^t  = 0.5, \lambda_{tri}^t =1.0,\lambda_{dis}^t= 0.5,\lambda_{sd}^t=1.0 $ and $\beta = \gamma = 0.5$. Similar to MMT\cite{ge2020mutual_iclr2020}, we set the temporal momentum to 0.999. The learning rate is fixed at 0.00035 for over 60 epochs. Compared with the random erasing probability, we set the adaptive erasing probability to 0.4, and use K-means  algorithm to cluster the feature vectors. For  Market-1501 and DukeMTMC-reID datasets, we set the number of clustering as $500$, $700$, and $900$; for MSMT17, we set it to $500$, $1000$,  $1500$, $2000$. We test different clustering numbers since the target identity number is unknown.
 

\subsection{Comparison with State-of-the-Arts}
\label{subsec:sotacomparison}

\textbf{Transfer from a small dataset to a large dataset.} In Table \ref{tab:comparison_small_to_large}, we test our method by transferring from small dataset to large dataset. Compared to DukeMTMC and Market1501, MSMT17 contains a much larger number of images with more people. This transfer setting is much more challenging and closer to the real cases. We compare our proposed method with prior state-or-the-arts methods on two transfer settings, \textit{i.e.}, Market1501→MSMT17, DukeMTMC→MSMT17. In both settings, our method achieves the best performance. Comparing to prior works, our method achieves significant performance improvement on both top-1 rate and mAP score, \textit{e.g.}, $32.9$ (HLI) \textit{v.s.} $26.5$ (GLT \cite{Zheng_2021_CVPR2021}), Market1501→MSMT17, $34.8$ (HLI) \textit{v.s.} $26.5$ (ABMT \cite{chen2021enhancing_wacv2021}), DukeMTMC→MSMT17. The impressive performance improvement indicates that the three explored aspects in HLI benefit the network training from a unified perspective, even in a challenging scenario where the target domain has more images with more people.

\textbf{Transfer between datasets with similar sizes.} In Table \ref{tab:comparison}, we make performance comparison between our proposed HLI and prior state-or-the-arts methods \footnote{We can not compare with IDM (CVPR2021) \cite{dai2021idm_iccv2021} since it chooses a much stronger baseline than ours, \textit{e.g.}, a baseline of $77.0\%$ mAP, DukeMTMC→Market1501.} on transferring between two datasets with similar sizes, \textit{i.e.}, DukeMTMC→Market1501, and Market1501→DukeMTMC. Both CMC and mAP scores are listed in the table. When transferring from DukeMTMC to Market1501, our method achieves a competitive performance with \textbf{76.5\%} top-1 rate and \textbf{91.6\%} mAP score; when transferring from Market1501 to DukeMTMC, our method achieves comparable performance with \textbf{82.5\%} top-1 and \textbf{68.8\%} mAP.  Comparing to the latest works, such as NRMT \cite{zhao2020unsupervised_eccv2020}, MEB-Net \cite{zhai2020multiple_eccv2020}, we can find that our method outperforms them by a large margin, \textit{e.g.}, $76.5$ (HLI) \textit{v.s.} $71.7$ (NRMT \cite{zhao2020unsupervised_eccv2020}), DukeMTMC→Market1501;  $68.8$ (HLI) \textit{v.s.} $66.1$ (MEB-Net \cite{zhai2020multiple_eccv2020}), Market1501→DukeMTMC. The significant performance improvement, in our point of view, comes from that HLI considers more than one aspect in human learning imitation and make them collaborate in a unified manner. For example, MMT \cite{ge2020mutual_iclr2020}, NRMT \cite{zhao2020unsupervised_eccv2020}, MEB-Net \cite{zhai2020multiple_eccv2020} all mimic the human learning process but from only one aspect, \textit{e.g.}, MMT \cite{ge2020mutual_iclr2020}, MEB-Net \cite{zhai2020multiple_eccv2020} focus on providing better teacher, NRMT \cite{zhao2020unsupervised_eccv2020} proposes to select reliable learning materials from existing data. Comparing to those prior works, our method achieves more advancement as it considers various aspects in human learning process. When comparing to MMT \cite{ge2020mutual_iclr2020}, we notice that the performance improvement is more significant if we choose ResNet-50 as the backbone, $74.3\%$ \textit{v.s.} $71.2\%$. Our guess is that our proposed HLI, benefiting from imitating human learning from three aspects, compensates the low network capability (ResNet-50 \textit{v.s.} IBN-ResNet-50).

\textbf{Discussion.} It is observed that when transferring from small dataset to large dataset, the performance improvement achieved by our method is significant than previous state-of-the-art methods, \textit{i.e.}, HLI achieves the best performance on Market1501→MSMT17 and DukeMTMC→MSMT17. We think the reason is that, when transferring from small data spaces to large data spaces (target), it is difficult to capture the target distribution if only utilizing the original learning materials in small data spaces. As HLI is able to update the learning materials adaptively and make it collaborate with the other two components, HLI achieves significant performance improvement under this challenging circumstance, \textit{e.g.},  Market1501→MSMT17 and DukeMTMC→MSMT17.

\begin{table} 
\centering
\resizebox{\linewidth}{!}{ 
\begin{tabular}{l|cccc}
\hline
\multicolumn{1}{c}{\multirow{2}{*}{Domain Adaption}} & \multicolumn{4}{|c}{DukeMTMC→Market1501}  \\
\cline{2-5}
\multicolumn{1}{c|}{}                         & mAP      & top-1     & top-5     & top-10         \\
\hline
Baseline                                      &69.8      & 89.7     &95.8       &97.2\\
Baseline+ALMS               &72.4      &90.7      &96.4       &97.6\\
Baseline+ALMS+AULM &73.2   &90.4      &96.2       &98.0\\
HIL &73.9  &90.6   &96.6   &98.0\\
\hline
\end{tabular}
} 
\caption{
Evaluation of effectiveness of each component in HLI on DukeMTMC→Market1501.
} 
\label{tab:ablations}
\end{table}

\subsection{Ablation Studies}
\label{subsec:ablation}
To demonstrate the efficacy and contribution of each component of our proposed method, we conduct extensive ablation studies on DukeMTMC→Market1501.

\textbf{The Impact of Each Component.} We conduct an experiment to test the effect of each component in HIL, \textit{i.e.}, adaptive updating learning materials (AULM), imitating teacher behavior selectively (ITBS), and analyzing learning materials structurally (ALMS). The results are listed in Table \ref{tab:ablations}. From the table, we can find each component in HIL has a positive influence on the final performance. For example, by considering the structural knowledge between learning materials, the mAP score increases from $69.8\%$ to $72.4\%$; with updating learning materials adaptively, the mAP score is further improved by $0.8\%$; by combining all three components together into a unified teacher-student framework, the final performance is boosted to $73.9\%$.



\textbf{Performance Sensitivity w.r.t the Material Update Frequency } When conducting adaptive learning materials update, we follow an update frequency, which is denoted as $prob$. We test different values of $prob$, \textit{i.e.}, $0$, $0.3$, $0.5$, $0.7$. We conduct the experiment based on ResNet-50. The mAP score with respect to $prob$ is shown in Table. \ref{tab:probupdate}. From the figure, we can find that the best mAP ($73.2$) is achieved when we set $prob$ as $0.4$. We notice that the model achieves the lowest mAP score when we set the $prob$ as $0.7$. This observation demonstrates that updating learning materials too frequently might hinder the student learning, which matches the scenario in the real world.



\begin{table} 
\centering
{ 
\begin{tabular}{l|cccc}
\hline

Prob &0  &0.4  &0.5   &0.7\\
\hline
mAP &72.4  &73.2   &71.7   &71.3\\
\hline
\end{tabular}
} 
\caption{
Performance sensitivity with respect to the learning materials update frequency.
} 
\label{tab:probupdate}
\end{table}

\begin{figure}[t]
  \centering
   \includegraphics[width=0.8\linewidth]{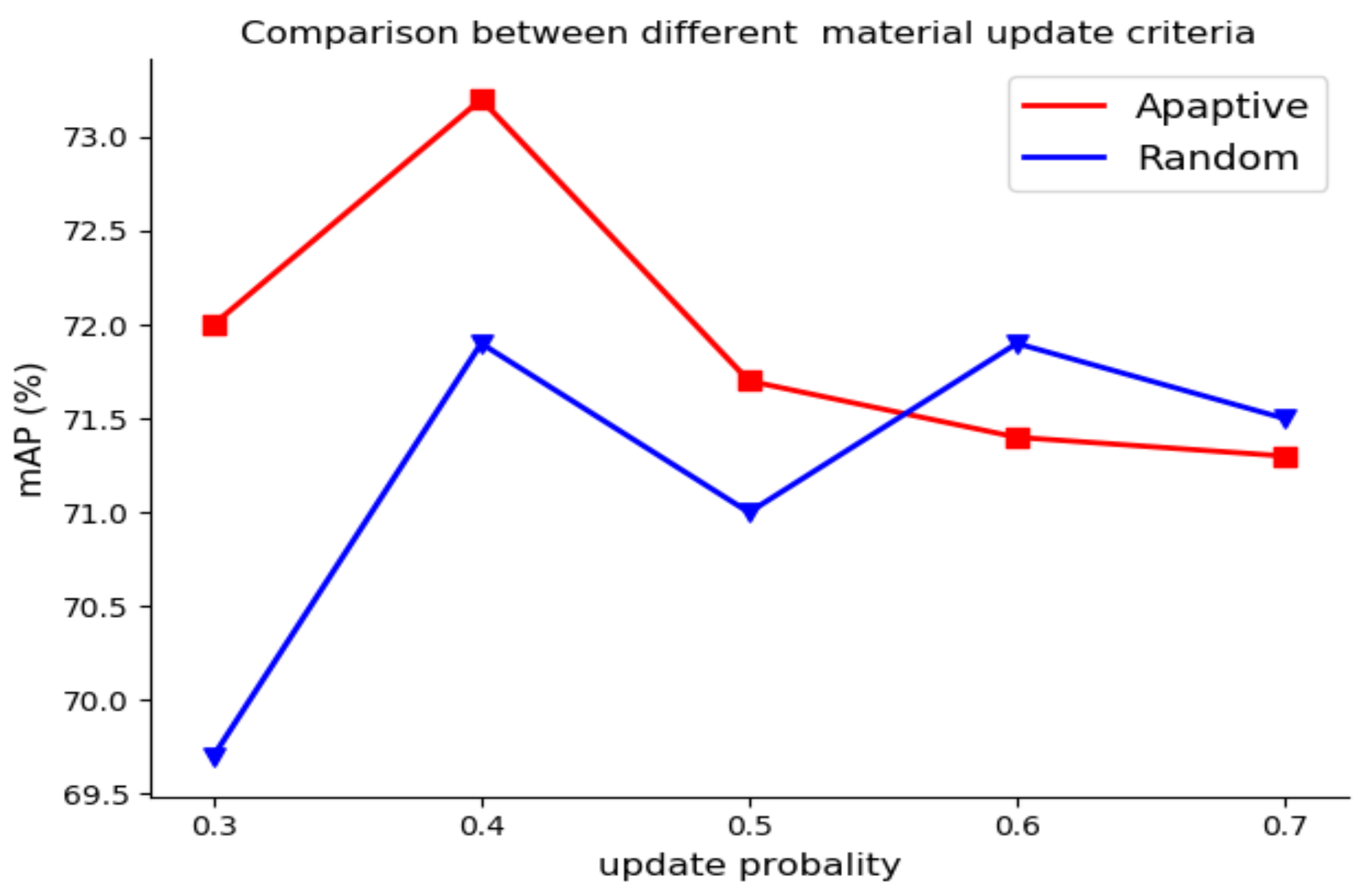}
\caption{Comparison between different material update criteria.}
   \label{fig:randomvsadaptive}
\end{figure}

\textbf{Performance Comparison between Random Update and Adaptive Update.} We conduct an experiment to demonstrate the necessity of updating the materials adaptively. We replace the adaptive update with a random update manner but with the same updating probability. We report the performance in Fig. \ref{fig:randomvsadaptive}, in which we can find that updating learning materials adaptively achieves the best performance and outperforms the random way in most cases. 

\begin{figure}[t]
  \centering
  \includegraphics[width=0.8\linewidth]{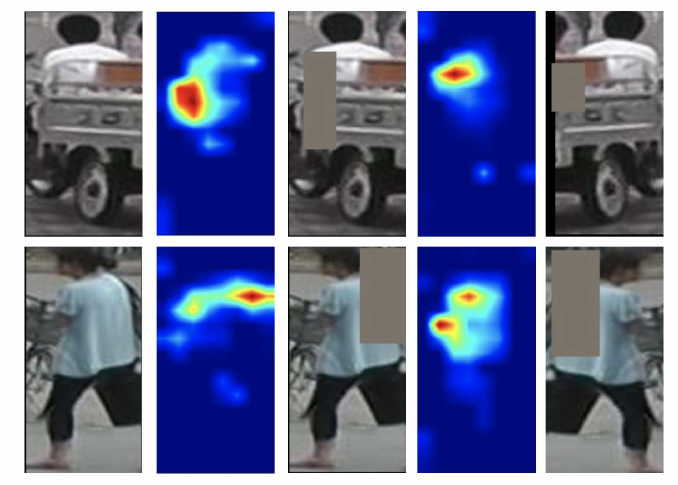}

  \caption{Visualization of Adaptively Updated Learning Materials. On one hand, for each given learning material (the first column), it collaborates with the network to generate updated learning materials, forcing the network to improve itself by mining various knowledge from the updated materials (the second and third columns). The updated network, on the other hand, collaborates with updated materials to generate new ones (the fourth and fifth columns). From left to right: original images $\boldsymbol{x}$, knowledgeable regions mined by a network $F(.; \theta)$, updated learning materials $\boldsymbol{x}'$ utilized to achieve an updated network $F'(.; \theta)$, knowledgeable region mined by the updated network $F'(.; \theta)$, updated learning materials $\boldsymbol{x}''$ utilized to achieve an updated network $F''(.; \theta)$. }
  \label{fig:visualizationheatmap}
\end{figure}

\begin{figure}[]
  \centering
  \includegraphics[width=0.9\linewidth]{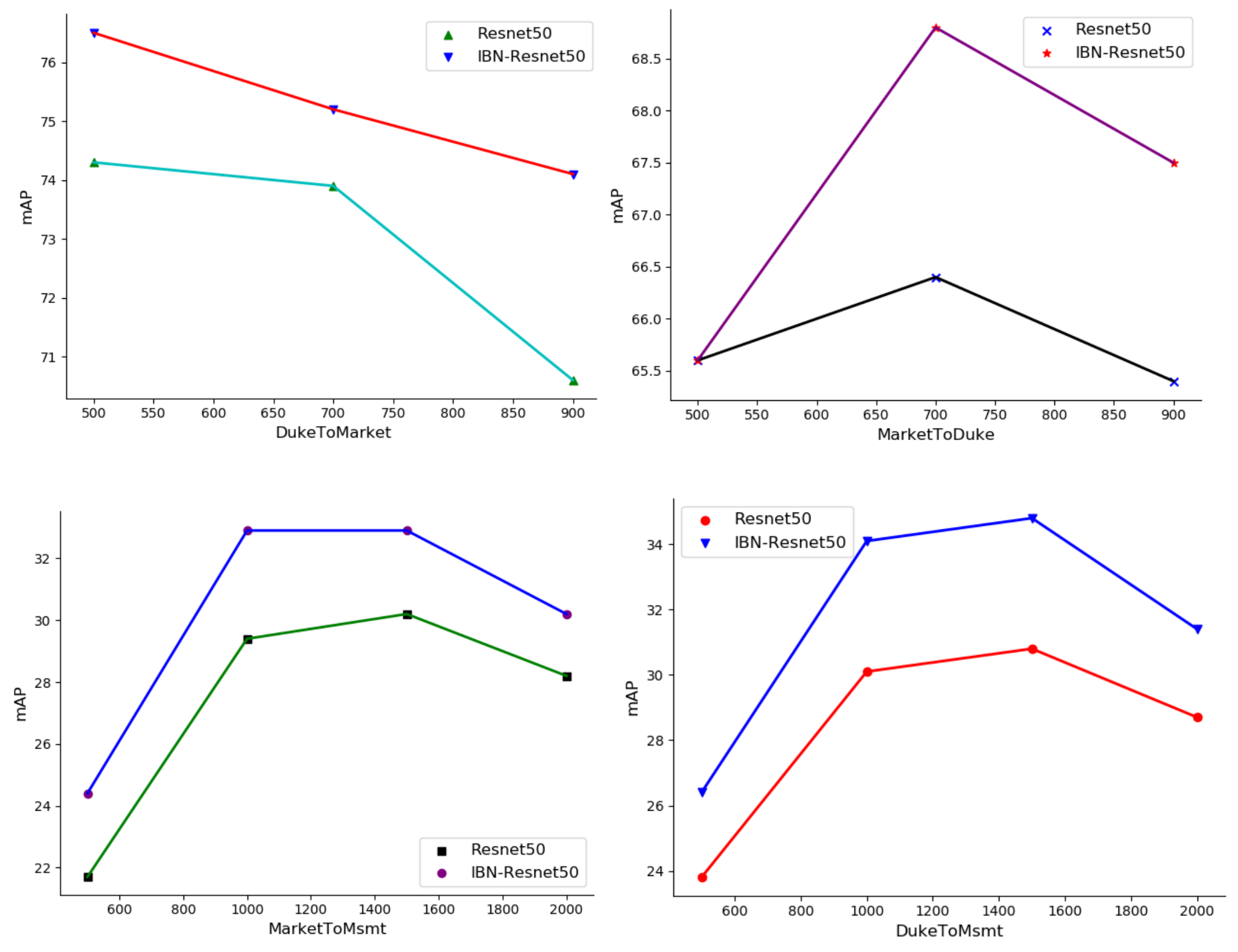}
\caption{Performance sensitivity with respect to clustering numbers.}
   \label{fig:parameter analysis1}
\end{figure}

\textbf{Performance sensitivity with respect to clustering numbers }
We set different clustering numbers for the performance sensitivity analysis with respect to clustering numbers. Specifically, for transferring from a small dataset to a larger dataset, \textit{i.e.}, from DukeMTMC to MSMT, from Market1501 to MSMT, we set the clutering number as $500$, $1000$, $1500$, $2000$; for transferring between datasets with a similar size, \textit{i.e.}, DukeMTMC, Market1501, we set the clustering number as $500$, $700$, and $900$. The experiment results are shown in Figure \ref{fig:parameter analysis1}. We can find that our method outperforms MMT, which is the most similar with our method, on all clustering numbers. This  demonstrates the superiority of HLI,  benefiting from the three aspects of HLI.

\begin{figure}[]
  \centering
   \includegraphics[width=0.9\linewidth]{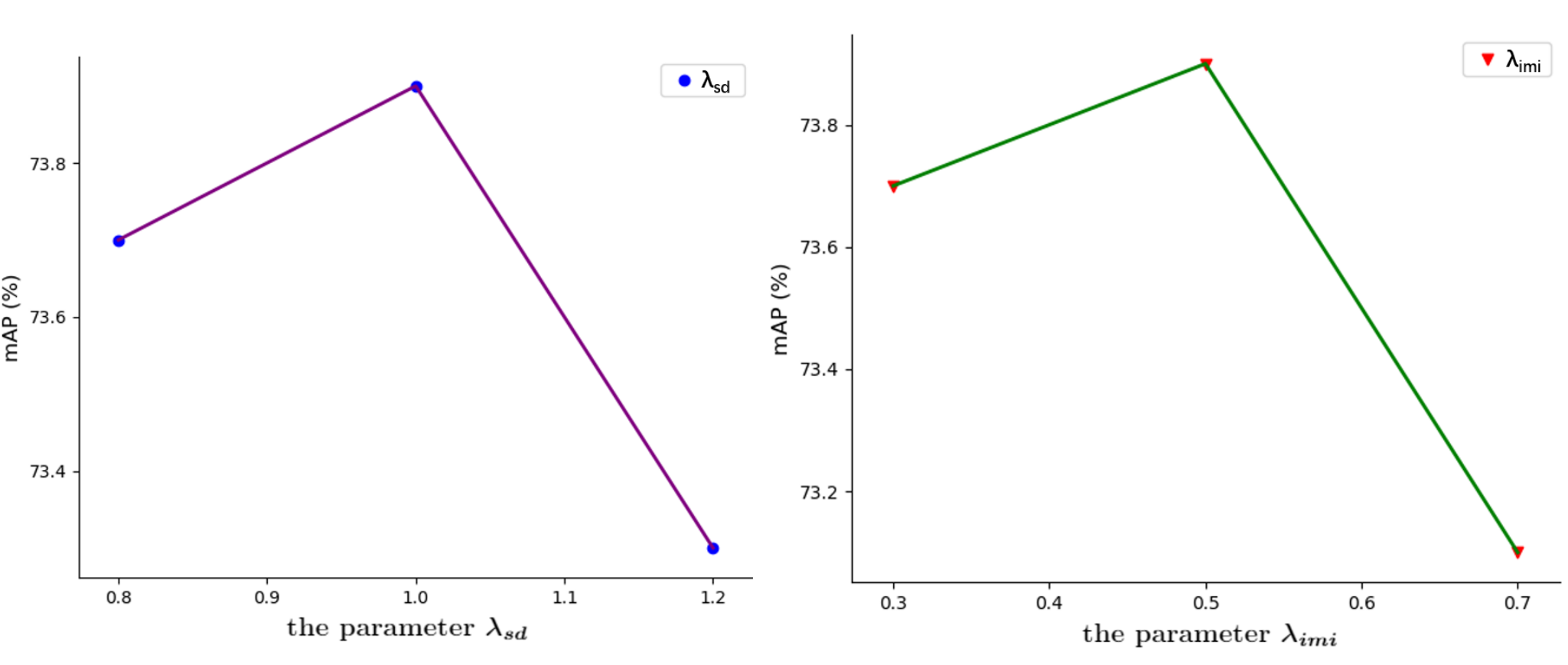}
\caption{Performance sensitivity with respect to weight parameters.}
   \label{fig:parameter analysis}
\end{figure}

\textbf{Performance sensitivity with respect to weight parameters }
We conduct ablation studies on loss functions with different weight parameters $\lambda_{imi}^t$ and $\lambda_{sd}^t$. We make the experiment on the setting of Duke-to-Market with a ResNet-50 backbone and use K-means algorithm with clustering number 700. We change the value of $\lambda_{sd}^{t}$ from $0.8$ to $1.2$, with a step size of $0.2$. For $\lambda_{imi}^{t}$, we set the value from $0.3$ to $0.7$ with a step size of $0.2$. The values for other loss weight parameters are set as the same values used in MMT. The results are shown in Figure \ref{fig:parameter analysis}. From the figure, we can observe a slight performance fluctuation with the change of the weight parameters, which motivates us to explore efficient methods to locate the optimal parameter configurations in our future work. 

\subsection{Result Visualization}
We show the adaptive updated learning materials in Fig. \ref{fig:visualizationheatmap}. Given the images, as the network updates its parameters, our method adaptively generates new learning materials by considering the network status and image content. As shown in the figure, even for the same images, the updated network will generate different updated new learning samples.



\section{Conclusions}
 In this work, we conduct explorations and find new ways to imitate the human learning process, \textit{i.e.}, adaptively updating learning materials, selectively imitating teacher behaviors, and analyzing learning materials structures. The explored components collaborate together to constitute a new method, \textit{i.e.}, HLI, for unsupervised domain adaptive person re-identification. The experimental results on three benchmark datasets demonstrate the efficacy of HLI.


%





\ifCLASSOPTIONcaptionsoff
  \newpage
\fi

\bibliographystyle{IEEEtran}
\bibliography{egbib}

\end{document}